\documentclass[11pt]{article}

\usepackage[final]{acl}

\usepackage{times}
\usepackage{latexsym}

\usepackage[T1]{fontenc}

\usepackage[utf8]{inputenc}

\usepackage{microtype}

\usepackage{inconsolata}

\usepackage{graphicx}

\usepackage{multirow}
\usepackage{array}
\usepackage{color}
\usepackage{tabularx}
\newcolumntype{Y}{>{\centering\arraybackslash}X}
\usepackage{wrapfig}
\usepackage{caption}
\usepackage{booktabs}
\usepackage{comment}
\usepackage{float}
\usepackage{stfloats}
\usepackage{subcaption}

\title{BiNSGPS: Geometry Problem Solving via Bidirectional Neuro-Symbolic Interaction}

\author{
    Qi Wang$^{1,2}$\thanks{~~~Equal contribution.},
    Peijie Wang$^{1,2}$\footnotemark[1],
    Fei Yin$^{1,2}$,
    Cheng-Lin Liu$^{1,2}$\thanks{~~~Corresponding author.}\\
    $^1$MAIS, Institute of Automation of Chinese Academy of Sciences\\
    $^2$School of Artificial Intelligence, University of Chinese Academy of Sciences\\
    \texttt{wangqi226@mails.ucas.ac.cn \quad wangpeijie2023@ia.ac.cn} \\
    \texttt{\{fyin, liucl\}@nlpr.ia.ac.cn}\\
    \rule{0pt}{1.5cm}
}

\begin{document}
\maketitle
\begin{abstract}
Geometry problem solving poses distinct challenges in artificial intelligence. Existing approaches typically fall into two paradigms: symbolic methods, which exhibit limited adaptability, and neural methods, which are prone to hallucinations. Recent neuro-symbolic hybrids predominantly rely on a unidirectional pipeline where neural outputs are fed into solvers without feedback, making system brittle to early-stage errors. To break this unidirectional bottleneck, we propose BiNSGPS, a framework that establishes Bidirectional Neuro-Symbolic Interaction (BiNS) between a MLLM Adviser and a Symbolic Solver. MLLM Adviser actively incorporates feedback from the symbolic solver to dynamically rectify inconsistent formal representations or propose auxiliary hypotheses, resolving symbolic conflicts and facilitating complex deductions.
Experiments show that BiNSGPS achieves state-of-the-art performance 
, reaching 90.5\% completion accuracy on Geometry3K and 90.1\% on PGPS9K, 
while maintaining 96\% step-wise logical coherence to balance capability with mathematical rigor.
\end{abstract}

\begin{figure*}[tb]
  \centering
  \includegraphics[width=\linewidth]{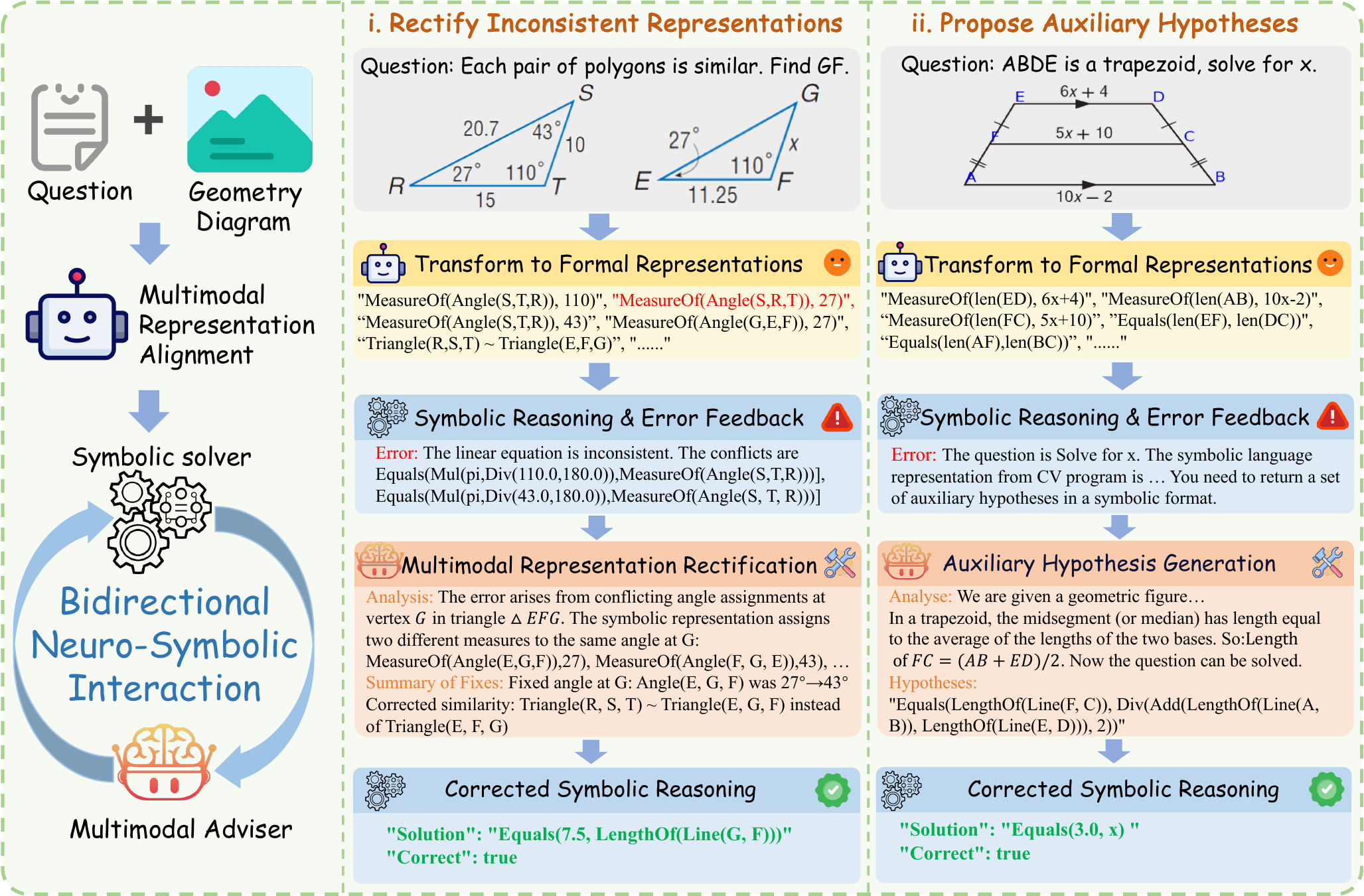}
  \caption{\textbf{Overview of our BiNSGPS.} 
  BiNSGPS processes multimodal inputs (text and diagram) through an iterative reasoning loop. Upon detecting inconsistencies (Case i: conflicting angle measures) or missing constraints (Case ii: trapezoid properties), the MLLM Adviser triggers representation rectification or auxiliary hypothesis generation. The two examples on the right illustrate how the closed-loop feedback transforms initial attempts (marked with \textcolor{red}{red warnings}) into verified solutions (\textcolor{green}{green checks}).}
  \label{fig:overview}
\vspace{-0.5cm}
\end{figure*}

\vspace{-0.5cm}
\section{Introduction}
\vspace{-0.1cm}
Geometry Problem Solving (GPS) stands as a quintessential challenge in artificial intelligence, serving as a rigorous task for evaluating high-level cognitive synthesis~\cite{AlphaGeometry,solidgeo,zhang2024proposingsolvingolympiadgeometry}. 
Mainstream GPS approaches traditionally diverge into two paradigms: \textbf{symbolic methods} and \textbf{neural methods}~\cite{PGPSnet,lans, zhang2024fusereasonverifygeometry}. Symbolic methods offer high precision and logical rigor~\cite{intergps,formalgeo,formalgeo2}, yet they are hampered by a significant formalization bottleneck. Translating raw problems into structured representations heavily depends on human annotations, and since symbolic methods are highly sensitive to initial parsing errors, incorrect automated formalization will lead to system collapse~\cite{NEURIPS2022_d0c6bc64, Autoformalsyntax,autoformalize}. Furthermore, the reasoning scope of symbolic systems is inherently restricted by predefined theorem rule sets, limiting their adaptability to diverse geometric configurations~\cite{trustgeogen,autogps}. Conversely, neural methods—ranging from specialized networks to Multimodal Large Language Models (MLLMs)—demonstrate superior flexibility by directly ingesting multimodal inputs~\cite{lans,gllava,mathllava,chen2025recoverable,miniinternvl,internvl3,mvmath}. However, these models are notoriously susceptible to \textbf{hallucinations}; they often generate superficially plausible yet logically flawed deductions, which severely undermines their mathematical credibility.

Recently, neuro-symbolic methods have emerged as a promising direction to synergize the multimodal comprehension of neural models with the rigorous deduction of symbolic engines~\cite{nssurvey,chen2026an}. However, current frameworks predominantly rely on a unidirectional pipeline, where neural models serve as front-end parsers to translate multimodal inputs into formal representations for a back-end symbolic solver~\cite{autogps}. This architecture creates a critical vulnerability: the information flow is strictly forward-moving, leaving the symbolic engine as a passive recipient of potentially flawed inputs. In complex geometric scenarios, neural formalization is notoriously prone to perceptual errors, while symbolic engines are often constrained by the inherent incompleteness of predefined theorem sets. Without a bidirectional feedback mechanism to "talk back" from the reasoning stage, the system lacks the agency to either rectify parsing hallucinations or leverage neural intuition to augment reasoning when the symbolic solver reaches a deadlock. Consequently, even a minor initial representation error or a missing theorem can lead to the failure of the entire reasoning chain.

To address these bottlenecks, we propose \textbf{BiNSGPS}, a novel framework that reformulates the neuro-symbolic paradigm as an MLLM-based agentic tool-calling architecture. Central to our approach is the \textbf{Bi}directional \textbf{N}euro-\textbf{S}ymbolic Interaction (\textbf{BiNS}), which breaks the traditional unidirectional constraint by establishing an active feedback loop between the MLLM Adviser and the Symbolic Solver, thereby enabling a bidirectional flow of information. Within this closed-loop system, the Symbolic Solver performs rigorous deduction via hypergraph expansion, identifying logical inconsistencies or deductive deadlocks. Specifically, BiNS enables the Symbolic Solver to generate diagnostic feedback that triggers the MLLM Adviser to either rectify inconsistent formal representations or propose auxiliary hypotheses derived from the MLLM’s internal knowledge, effectively circumventing the limitations of predefined theorem rule sets. As illustrated in Fig.~\ref{fig:overview}, the MLLM Adviser can dynamically correct perceptual errors identified by the Symbolic Solver or introduce external geometric intuition to break reasoning deadlocks. Comprehensive experiments demonstrate that BiNSGPS achieves state-of-the-art (SOTA) accuracies while effectively balancing mathematical rigor with reasoning flexibility.

Our contributions are summarized as follows:

\textbf{1.Interactive Correction and Augmentation:} We introduce \textbf{BiNS} mechanism into neural-symbolic methods, where the Symbolic Solver provides diagnostic feedback to MLLM Adviser. This interaction enables rectification of formalization errors and heuristic augmentation of reasoning.

\textbf{2.Bidirectional Neuro-Symbolic Framework:} We propose \textbf{BiNSGPS}, reformulating the traditional unidirectional neuro-symbolic pipeline into an agentic, tool-calling architecture. By establishing a feedback loop between the MLLM Adviser and the Symbolic Solver, we break the rigid information bottleneck inherent in conventional GPS.

\textbf{3.SOTA Performance:} Extensive evaluations on Geometry3K and PGPS9K demonstrate that BiNSGPS achieves new SOTA performance, reaching 90.5\% and 90.1\% solving accuracy. BiNSGPS also ensures exceptional logical consistency in step-wise deductions, establishing a superior balance between reasoning flexibility and mathematical rigor.

\section{Related Work}
\vspace{-0.2cm}
\subsection{Geometry Problem Solving}
\vspace{-0.1cm}
Geometry Problem Solving (GPS) is a hallmark of mathematical intelligence. Existing literature can be broadly categorized into three paradigms: 
1. Neural-based methods treat GPS as a visual question answering task. Early works utilized specialized architectures like PGPSNet~\cite{pgdpnet} and LANS~\cite{lans}, while recent efforts leverage the reasoning power of MLLMs such as G-LLaVA~\cite{gllava} and Vision-R1~\cite{vision-r1}. While these models excel at flexible intuition, they are prone to hallucinations and lack the rigor required for mathematical proofs. 
2. Symbolic-based methods (e.g., InterGPS~\cite{intergps}, E-GPS~\cite{10654817}) translate raw inputs into structured formal languages and execute deduction. Although logically rigorous, these systems suffer from a formalization bottleneck: they are highly sensitive to initial parsing errors and lack the adaptability to handle configurations outside their hard-coded theorems. 
3. Neuro-symbolic methods attempt to bridge this gap. Recent breakthroughs like AlphaGeometry~\cite{AlphaGeometry} and AlphaGeometry2~\cite{AlphaGeometry2} combine language models with symbolic engines to solve Olympiad problems. Other approaches like AutoGPS~\cite{autogps} and GeoDRL~\cite{geodrl} use MLLM components to guide theorem selection or generate aligned representations. Geoparsing~\cite{wang2026geoparsing} further advances the perception stage by introducing a unified formal language for both plane and solid geometry diagram parsing, enabling MLLMs to convert geometric diagrams into structured symbolic descriptions that can serve as cognitive scaffolds for downstream reasoning. However, these systems predominantly rely on a unidirectional pipeline where the Symbolic Solver is a passive recipient of neural network' outputs. This one-way flow means any parsing error is irrecoverable and the solver cannot "talk back" to the neural component for help.
In contrast, our BiNSGPS framework introduces a bidirectional interaction. By establishing a feedback mechanism, the Symbolic Solver can prompt the MLLM to rectify inconsistent representations and hypothesize auxiliary constructions, breaking the "one-way" bottleneck of previous neuro-symbolic methods.

\vspace{-0.2cm}
\subsection{MLLMs for Geometry Reasoning}
\vspace{-0.1cm}
Multimodal Large Language Models (MLLMs) have become a focal point for mathematical reasoning, with specialized benchmarks such as MathVista~\cite{mathvista}, MATH-Vision~\cite{mathvision}, GeoEval~\cite{geoeval}, MV-MATH~\cite{mvmath}, SolidGeo~\cite{solidgeo} and GeoLaux~\cite{fu2025geolauxbenchmarkevaluatingmllms}, highlighting their potential. Research in this area primarily follows two parallel trajectories. One direction focuses on Supervised Fine-Tuning and Reinforcement Learning~\cite{interactiveevolution, symbolllm, deepseekmath, deepseekmathv2} to adapt models to the geometry domain, resulting in models like G-LLaVA~\cite{gllava} and GeoGPT4V~\cite{geogpt4v}. While these models exhibit improved intuition, they remain susceptible to both perceptual errors in diagram parsing and logical hallucinations during the deduction process.
Another research trajectory treats MLLMs as agents that interact with external tools to solve complex tasks~\cite{ToRA, Suris_2023_ICCV, pan-etal-2023-logic}. Frameworks such as Visual SKETCHPAD~\cite{NEURIPS2024_fb820110} and CODEPLOT-COT~\cite{codeplot} exemplify this approach, utilizing auxiliary drawing modules or code-execution environments to assist the MLLM in visual analysis and calculation.

In this work, we follow the agent-based paradigm by adopting a tool-calling architecture. Within the BiNSGPS framework, MLLM acts as a central coordinator that interfaces with specialized neural network and symbolic solvers. This approach allows the MLLM to leverage the high-fidelity outputs of domain-specific tools while focusing on strategy and error correction.

\vspace{-0.2cm}
\section{Methodology}

\begin{figure*}[th]
  \centering
  \includegraphics[width=\linewidth]{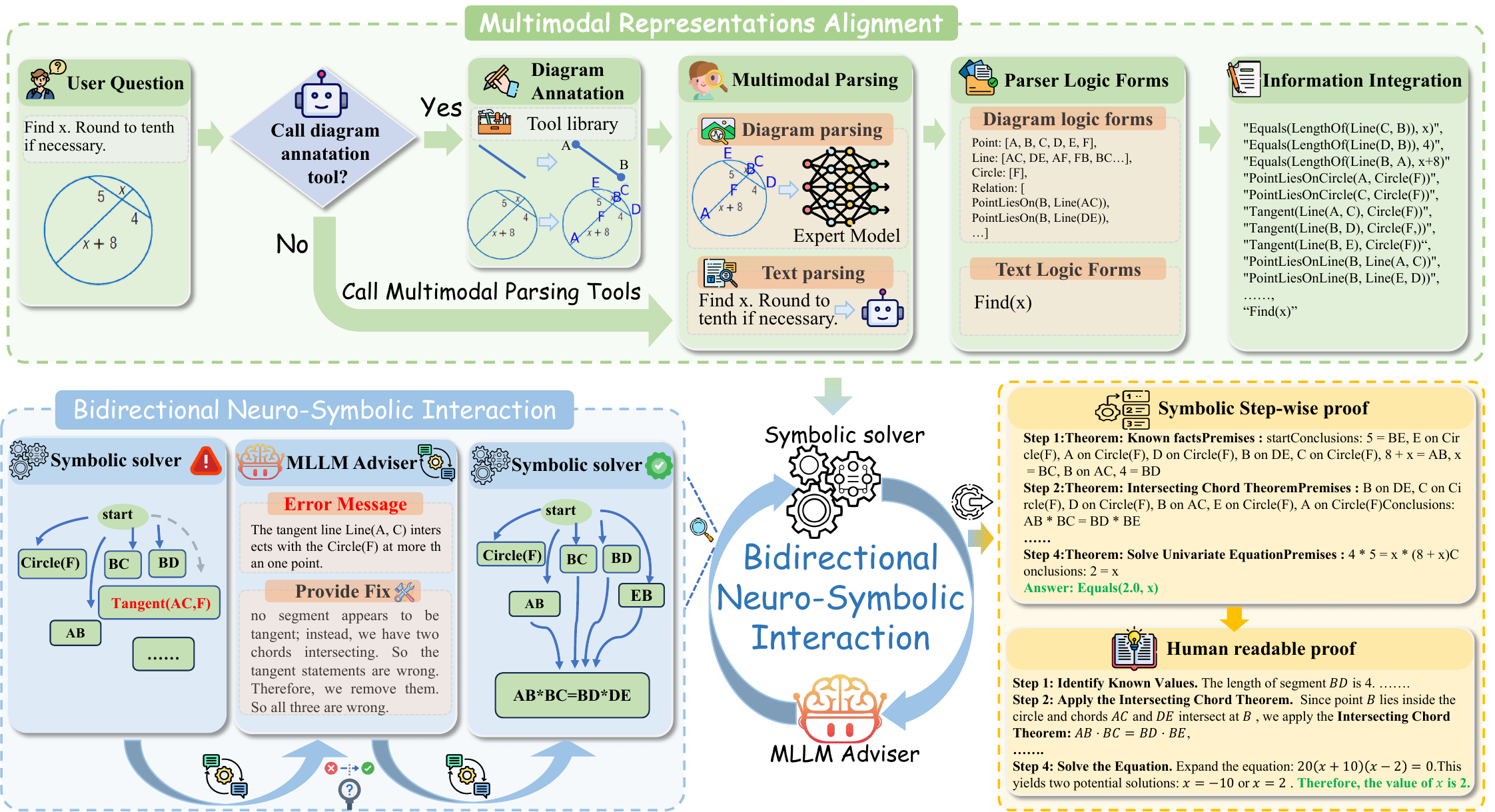}
  \caption{Framework of BiNSGPS pipeline. \textbf{Top}: Multimodal Representations Alignment. Geometry diagram-text pairs first get annotation. Then the pair are processed through a dual-parser architecture where a specialist neural network extracts structural diagram primitives while an MLLM parses textual constraints. These are integrated into initial logic forms $\mathcal{L}$, which are aligned and completed to form a comprehensive symbolic representation.
  \textbf{Bottom}: Bidirectional Neuro-Symbolic Interaction \& Proof Generation. The Symbolic Solver constructs hypergraph for deductive reasoning. Upon encountering inconsistencies or deadlocks, the MLLM Adviser analyzes error diagnostics to rectify erroneous formalizations or propose auxiliary hypotheses $\mathcal{H}$. Once a solution is reached, the Solver identifies the minimal reasoning subgraph, which the MLLM subsequently translates into a human-readable proof.}
  \label{fig:pipeline}
\vspace{-0.5cm}
\end{figure*}

\vspace{-0.2cm}
\subsection{Overview}
The BiNSGPS framework is illustrated in Fig.~\ref{fig:pipeline}. Our BiNSGPS framework is a closed-loop neuro-symbolic system comprising three primary components: Multimodal Representations Alignment, Symbolic Solver, and MLLM Adviser. The system follows an iterative execution flow to bridge the gap between neural intuition and symbolic rigor. Given a geometry problem ($\mathcal{D}$, $\mathcal{T}$), the diagram $\mathcal{D}$ and text $\mathcal{T}$ are first processed by the Multimodal Representations Alignment module to generate a formal specification in the form of an initial representations set $\mathcal{L}$ = ($l_1$, $l_2$, ..., $l_n$), which serves as the foundational geometric facts for the problem.

The Symbolic Solver then ingests $\mathcal{L}$ to perform deduction. Equipped with a predefined theorem rule set $R$, the solver formulates the reasoning process as a hypergraph expansion task.
If the solving fails (including deductive deadlock or logical contradictions), it triggers the Bidirectional Feedback mechanism. The Symbolic Solver summarizes the current state and communicates it to the MLLM Adviser. In the case of conflict, the Adviser identifies and rectifies the inconsistent elements within $\mathcal{L}$ based on the solver's error trace to resolve inconsistency.If the solver reaches a deadlock, the Adviser analyzes the state to propose a hypothesis set $\mathcal{H}$. These auxiliary representations are provided as a supplement to $\mathcal{L}$, enabling the Symbolic Solver to bypass the "limited theorem" bottleneck.

This bidirectional loop continues until a verifiable solution is found or the maximum iteration limit is reached. If problem solved, the Symbolic Solver extract a minimal solution path, and then the MLLM polishes the solution into a human-readable format. If reach the iteration limit, the system employs a fallback mode, where the MLLM generates a direct numerical answer based on the existing representations. The details of each component are elaborated in the following sections.

\vspace{-0.2cm}
\subsection{Multimodal Representations Alignment}
\vspace{-0.1cm}
The Multimodal Representations Alignment module transforms diagrams $\mathcal{D}$ and text $\mathcal{T}$ into a unified set of formal symbolic representations. This process is carried out by the MLLM through tool invocation across three stages: Standardized Annotation, Dual-Source Extraction, and Unified Information Integration.

In the standardized annotation stage, the diagram is standardized for precise geometric extraction. The MLLM first evaluates the diagram to identify critical unlabeled features. If necessary, it invokes a specialized small-scale neural model to label points with unique letter identifiers. This results in an annotated diagram $\mathcal{D'}$, which ensures that all subsequent symbolic references are consistent across the vision and text modules.

In the dual-source extraction stage, the system extracts geometric facts from both visual and textual sources. The MLLM invokes PGDPNet to parse $\mathcal{D'}$. By leveraging PGDPNet’s superior accuracy ($>99\%$) in element and position relation recognition, the system compensates for the inherent perceptual fragility of general-purpose MLLMs. This sub-module extracts basic elements $\mathcal{B}$ (points, lines, circles), positional relations $\mathcal{P}$ (e.g., point on line) and other geometry representations $\mathcal{F}$. For the textual extraction, the MLLM acts as a text parser via few-shot prompting to extract textual assertions $\mathcal{T'}$. To resolve references like "quadrilateral in the figure," the MLLM is provided with $\mathcal{D'}$ to align textual descriptions with visual grounding. The combined output is structured into a standardized JSON, encompassing $\mathcal{B}$, $\mathcal{P}$, additional formal assertions $\mathcal{F}$, and processed text $\mathcal{T'}$.

In information integration and transformation stage, The MLLM performs a holistic analysis of the extracted components to synthesize the final representation set $\mathcal{L}$. Crucially, the MLLM is instructed to prioritize the high-fidelity visual facts ($\mathcal{B}$ and $\mathcal{P}$) from the diagram parser to calibrate and rectify potential errors in the textual extraction $\mathcal{T'}$.

\vspace{-0.2cm}
\subsection{Symbolic Solver}
\vspace{-0.1cm}
Upon receiving the initial representation set $\mathcal{L}$ from the alignment module, the Symbolic Solver initiates a structured deduction process. Symbolic Solver formulate GPS as a Hypergraph Construction task, where geometric facts are represented as nodes and theorems as hyperedges. The solving procedure is executed in two primary stages: Geometric Completion and Iterative Graph Expansion.

Before formal deduction, the solver performs a completion step to ensure the initial representation set is logically complete and standardized. This stage automates the derivation of implicit properties from given definitions to ensure that subsequent reasoning is not stalled by missing foundational facts. After completion, the solver proceeds to the graph construction stage.

In graph expansion stage, the initial representations $\mathcal{L}$ are treated as root nodes. The reasoning process is governed by a comprehensive theorem library $\mathcal{R}$, which includes geometric rules and algebraic rules. For each derivation, the solver identifies a minimal feasible set of nodes $m \subset \mathcal{L}'$ that satisfies the antecedents of a theorem rule $r \in \mathcal{R}$. The conclusion $\mathcal{C}$ is then integrated into the graph as a new node. To ensure logical traceability and prevent circular reasoning, each node $\mathcal{C}$ maintains a directed pointer to its parent nodes $m$ and the specific rule $r$ applied.

Throughout the process, the solver maintains strict logical integrity. If the solver detects a conflict, it immediately halts the derivation to prevent the propagation of errors. This terminal failure state, along with cases of deductive deadlock where no new nodes can be generated, triggers the transition to the Bidirectional Feedback phase to resolve the underlying representational issues.

\vspace{-0.2cm}
\subsection{Bidirectional Feedback \& MLLM Adviser}
\vspace{-0.1cm}
The MLLM Adviser serves as the sophisticated cognitive coordinator of the BiNSGPS framework, transforming the traditionally passive role of MLLMs into an active, agentic component capable of responding to the rigorous demands of Symbolic Solver. By establishing a closed-loop interaction, the Adviser addresses the "unidirectional bottleneck" that renders conventional neuro-symbolic systems brittle to early-stage errors. This bidirectional communication allows the system to operate in two distinct strategic modes—Rectify Inconsistent Representations and Propose Auxiliary Hypotheses—each triggered by specific failure states identified during the Symbolic Solver’s solving process.

In rectify inconsistent representations mode, which addresses logical contradictions, MLLM Adviser functions as a diagnostic and corrective agent. When the Symbolic Solver identifies an inconsistency, the conflict predication, including the annotated diagram $\mathcal{D'}$, text $\mathcal{T}$, and the conflicting initial representation set $\mathcal{L}$, is passed to the MLLM Adviser. To provide a stable foundation for this correction, we leverage the high-fidelity output of the PGDPNet diagram parser, designating the basic geometric elements $\mathcal{B}$ and their positional relations $\mathcal{P}$ as anchored ground truths. MLLM uses them as an objective reference to calibrate and refine more ambiguous assertions. In this rectification phase, the Adviser is strategically constrained to the modification or deletion of existing representations in $\mathcal{L}$, ensuring that the resubmitted representation set is a more accurate reflection of the raw geometric input rather than a hallucinated expansion.

When the Symbolic Solver reaches deductive deadlock—a state where no further theorems from the library $\mathcal{R}$ can be applied—the Adviser transitions into Propose Auxiliary Hypotheses. This mechanism addresses the inherent limitation of symbolic systems, which are strictly bound by the scope of their predefined rule sets. Conversely, the MLLM can draw upon an expansive repository of geometric and algebraic knowledge acquired during its extensive pretraining, allowing it to propose novel conclusions $\mathcal{H}$ that serve as vital catalysts for the solver. In this mode, the system prioritizes problem-solving by prompting the MLLM to generate supplementary representations that bridge the gap between known facts and the target answer. To mitigate the risk of creative hallucinations, we enforce a Rationale-Conclusion sequence; the MLLM must articulate an analytical reasoning chain before presenting any new formal representation. If the Symbolic Solver subsequently detects a conflict within this new hypothesis set $\mathcal{H}$, the error is immediately fed back to the Adviser for iterative refinement, creating a self-correcting cycle that persists until reaching a consistent and solvable state.

By synergizing the logical rigor of hypergraph deduction with the adaptive intuition of MLLM, the Adviser creates a new paradigm of deep integration, enabling the system to transcend the limitations of neural-based and symbolic-based approaches.

\vspace{-0.2cm}
\subsection{Proof Generation}
\vspace{-0.1cm}
The final stage of the BiNSGPS ensures that the derived results are both mathematically verifiable and human-readable. This process is bifurcated based on the outcome of the iterative reasoning loop: either generating a refined proof from the symbolic trace or executing a fallback procedure.

Upon the successful derivation of the target value, the Symbolic Solver performs a recursive back-trace to find the minimal solution graph, details can be found in the Appendix.\ref{appendix:Symbolic Solver}.
The initial format is not suitable for human. To bridge this gap, the MLLM acts as a translator, converting the symbolic nodes and theorem applications into a more readable proof $S'$. This ensures that the final output maintains high logical alignment with the symbolic engine while friendly to users.

When the bidirectional interaction reaches the maximum iteration threshold, the system enters the fallback mode.
In this mode, the MLLM is provided with the textual description $\mathcal{T}$, the annotated diagram $\mathcal{D'}$, and the high-accuracy basic geometric elements $\mathcal{B}$ and positional relations $\mathcal{P}$. The MLLM is then prompted to provide the final output in a structured Analyze-Answer-Proof format. 
This ensures that when the solver cannot find a solution, the Adviser can provide an informed estimate supported by the structured visual grounding.

\vspace{-0.3cm}
\section{Experiments}
\vspace{-0.2cm}

To evaluate the effectiveness of BiNSGPS, we utilize two primary benchmarks in the geometry reasoning field: Geometry3K~\cite{intergps} and PGPS9K~\cite{PGPSnet}, which provide 3,001 and 9,022 geometry problem-image pairs, respectively. Each problem is formatted with four candidate choices. We assess our framework under two distinct settings: Choice mode, where the model selects from the provided options, and Competition mode, where the system must derive the exact numerical answer independently.
We also introduce a human-evaluation focused on Step-wise Logical Coherence. This metric is used to quantify the impact of MLLM hallucinations on the reasoning process, proving that the generated proofs are not merely superficially plausible but are mathematically sound to the symbolic derivations. 

\begin{table*}[t]
\centering
\caption{Performance comparison among state-of-the-art geometry problem solvers.}
\label{tab:performance_comparison}
\begin{tabularx}{\textwidth}{@{}lYYYY@{}}
\toprule
\multirow{2}{*}{Method} & \multicolumn{2}{c}{Geometry3K} & \multicolumn{2}{c}{PGPS9K} \\
 & Choice & Completion & Choice & Completion \\
\midrule
\multicolumn{5}{c}{\textbf{MLLMs}} \\
InternVL2.5-78B~\cite{internvl3} & 60.9 & 36.1 & 51.3 & 28.1 \\
Vision-R1-7B~\cite{vision-r1} & 57.1 & 43.8 & 49.6 & 36.8 \\
GPT-4o~\cite{openai2024gpt4o} & 57.1 & 46.3 & 46.0 & 37.2 \\
Qwen3-VL-7B~\cite{qwen3vl} & 50.1 & 44.8 & 44.9 & 43.0 \\
Qwen2.5-VL-32B~\cite{qwen2025qwen25technicalreport} & 58.6 & 46.3 & 56.1 & 43.5 \\
InternVL3-78B~\cite{internvl3} & 74.5 & 57.4 & 61.1 & 48.9 \\
Qwen3-VL-Plus~\cite{qwen3vl} & 67.4 & 59.8 & 72.2 & 58.4 \\
GPT-5.2 & 84.5 & 77.9 & 78.1 & 72.5\\
\midrule
\multicolumn{5}{c}{\textbf{Neural Solvers}} \\
NGS~\cite{geoqa} & 58.8 & 35.3 & 46.1 & 34.1 \\
Geoformer~\cite{geoformer} & 59.3 & 36.8 & 47.3 & 35.6 \\
PGPSNet~\cite{PGPSnet} & 77.9 & 65.2 & 69.4 & 62.7 \\
LANS~\cite{lans} & 82.3 & 72.1 & 74.0 & 66.7 \\
\midrule
\multicolumn{5}{c}{\textbf{Symbolic Solvers}} \\
InterGPS~\cite{intergps} & 63.5 & 50.6 & 66.2 & 57.4 \\
E-GPS~\cite{10654817} & 67.9 & - & - & - \\
GeoDRL~\cite{geodrl} & 68.4 & - & - & - \\
\midrule
\multicolumn{5}{c}{\textbf{Neural-Symbolic Solvers}} \\
GNS (LLaVA-1.5-13B)~\cite{GNS} & 53.8 & - & - & - \\
Pi-GPS~\cite{pigps} & 77.8 & 70.6 & 69.8 & 61.4 \\
AutoGPS (Qwen3-VL-Plus)~\cite{autogps} & 81.1 & 74.8 & 81.8 & 75.7 \\
AutoGPS (GPT-4o)~\cite{autogps} & 81.6 & 75.4 & 81.5 & 75.3 \\
\textbf{BiNSGPS (Qwen3-VL-Plus)} & \textbf{95.2} & \textbf{90.5} & \textbf{92.7} & \textbf{90.1}\\
BiNSGPS (Qwen3-VL-32B) & 90.3 & 87.0 & 89.2 & 87.6\\
\bottomrule
\end{tabularx}
\vspace{-0.35cm}
\end{table*}

\vspace{-0.25cm}
\subsection{Comparison with State-of-the-art Solvers}
\vspace{-0.1cm}
As summarized in Table~\ref{tab:performance_comparison}, BiNSGPS achieves a decisive performance advantage over state-of-the-art neural, symbolic, and multimodal large language models (MLLMs). In Choice mode, it outperforms the next best approach by 10.7\% on Geometry3K and 10.9\% on PGPS9K, reaching an accuracy of 95.2\%. This is largely because BiNSGPS utilizes candidate choices as a verification signal: if the symbolic output mismatches the options, the MLLM Adviser is triggered to rectify proof trace errors, avoiding the random selection typical of traditional solvers. Notably, in the challenging Completion mode on Geometry3K, BiNSGPS achieves 90.5\% accuracy, significantly surpassing frontier models like GPT-5.2 (77.9\%).

Despite their rapid evolution, pure MLLMs like GPT-5.2 fail to breach the 80\% accuracy threshold. This stems from a fundamental limitation in geometric grounding rather than visual resolution. Standard MLLM vision encoders employ a ``semantic-first'' tokenization that sacrifices spatial granularity for global context, leading to ``perceptual drift'' and missed mathematical precision. In contrast, BiNSGPS bridges this gap by offloading structural recognition to PGDPNet, which extracts rigorous geometric primitives. By anchoring the MLLM Adviser in this high-fidelity symbolic foundation, BiNSGPS eliminates pure-neural hallucinations, allowing the MLLM to focus on high-level reasoning while the Symbolic Solver guarantees mathematically valid deductions.

\vspace{-0.2cm}
\subsubsection{Step-wise Logical Coherence}
\vspace{-0.1cm}
While MLLMs can often arrive at a correct final answer through heuristic "shortcuts," their internal reasoning is frequently marred by logical hallucinations. To assess the impact of the MLLM on problem solving within BiNSGPS, we conduct a human evaluation of step-wise correctness on 100 questions of Geometry3K. The accuracy of common MLLMs and BiNSGPS is reported in Table~\ref{tab:stepwise_accuracy and MRA}.

\begin{table}[h]
\centering
\caption{Step-wise Logical Coherence and Multimodal Representations Alignment}
\label{tab:stepwise_accuracy and MRA}
\resizebox{0.98\columnwidth}{!}{
\begin{tabular}{lcc}
\toprule
MLLM model & Step Accuracy & Alignment \\
\midrule
Qwen2.5-VL-32B & 66.0\% & -\\
Qwen3-VL-32B & - & 56.0\% \\
GPT-4o & 70.0\% & 60.0\%\\
Qwen3-VL-Plus & 78.0\% & 58.0\%\\
Ours (BiNSGPS) & 96.0\% & 80.0\% \\
\bottomrule
\end{tabular}
}
\vspace{-0.5cm}
\end{table}

The evaluation reveals a significant disparity in reasoning quality. Even advanced models like Qwen3-VL-Plus exhibit a "hallucination gap," with nearly 22\% of their successful solutions containing at least one logically invalid step. In contrast, BiNSGPS achieves a coherence rate of 96.0\%. This superior rigor is a direct result of our multi-layered validation architecture: the high-fidelity visual grounding of PGDPNet, the rigorous constraint-checking of the Symbolic Solver.

\vspace{-0.2cm}
\subsubsection{MathVista benchmark}
\vspace{-0.1cm}
To further evaluate BiNSGPS, we evaluated performance on the MathVista GPS part~\cite{mathvista} against models from the official leaderboard, and the result is shown in Figure~\ref{fig:mathvista and iteratioin}(a).

As shown in the figure, our BiNSGPS get 92.9, outperforming industry leaders like Kimi-k1.6 and Step R1-V. It is surpassed only by DreamPRM (95.7), which relies on a Process Reward Model. Notably, BiNSGPS is a tuning-free architecture, demonstrating that state-of-the-art reasoning can a be achieved through bidirectional interaction without the overhead of specialized training or the risk of degrading the MLLM's general capabilities.

\vspace{-0.25cm}
\subsection{Ablation Study}
\vspace{-0.1cm}
To analyze the effectiveness of each component within the BiNSGPS framework, we conduct a series of ablation studies. These studies focus on Multimodal Representation Alignment and the MLLM Adviser. Specifically, we investigate how the specialist neural network and the bidirectional feedback loop influence the final performance.

\vspace{-0.2cm}
\subsubsection{Multimodal Representation Alignment}
\vspace{-0.1cm}
We compare the accuracy and completeness of geometric information extracted by the MLLM equipped with PGDPNet against few-shot MLLM extraction. Evaluations are conducted on Qwen3-VL-32B, Qwen3-VL-Plus, GPT-4o and BiNSGPS on a randomly chosen 100 scale subset of Geometry3K dataset. The results are reported in Table~\ref{tab:stepwise_accuracy and MRA}. 

As shown in Table~\ref{tab:stepwise_accuracy and MRA}, general-purpose MLLMs frequently struggle to reach 60\% alignment accuracy. This is largely due to "semantic hallucinations," where the models overlook fine-grained geometric details or fail to capture the complete set of relations in diagrams.
In contrast, BiNSGPS achieves an alignment accuracy of 80.0\%. This significant gain is attributed to the using of PGDPNet. By giving the extraction of basic geometric elements and spatial relationships to a dedicated network, we transform visually dense diagrams into high-fidelity symbolic primitives. This decoupling allows the MLLM to focus instead on identifying high-level geometric properties. By providing the MLLM with an "anchored" foundation of basic elements, we effectively mitigate the "perceptual drift" typical of pure-neural approaches, ensuring that the initial representation set $\mathcal{L}$ is both comprehensive and mathematically grounded.

\begin{table}[ht]
\vspace{-0.3cm}
\centering
\caption{Ablation study of MLLM Adviser.}
\label{tab:ablation_mllm}
\begin{tabularx}{\columnwidth}{@{}lYY@{}}
\toprule
Method & Geometry3K & PGPS9K \\
\midrule
No MLLM Adviser & 73.3\% & 72.1\% \\
rectify-only & 80.4\% & 75.6\% \\
hypothesis-only & 77.6\% & 79.2\% \\
BiNSGPS & \textbf{90.5\%} & \textbf{90.1\%} \\
\bottomrule
\end{tabularx}
\vspace{-0.5cm}
\end{table}

\subsubsection{MLLM Adviser and Feedback}
To isolate the impact of the bidirectional feedback loop, we compare the full BiNSGPS framework against an ablated version lacking the MLLM Adviser. In this "No Adviser" configuration, the system functions as an unidirectional pipeline. We also provide the data about only keep one function of MLLM Adviser as "rectify-only" and "hypothesis-only". The results are summarized in Table~\ref{tab:ablation_mllm}.

As shown in the Table~\ref{tab:ablation_mllm}, the inclusion of the MLLM Adviser yields a transformative improvement in answer accuracy, with gains of 17.2\% on Geometry3K and 18.0\% on PGPS9K in Completion mode.
This performance surge is primarily driven by two synergistic functions: Iterative Error Recovery, where the MLLM uses diagnostic feedback to rectify initial parsing hallucinations; Knowledge Augmentation, which allows the system to bypass theorem-set limitations by proposing auxiliary hypotheses; 
By transforming the Symbolic Solver from a passive executor into an active diagnostic partner, BiNSGPS overcomes the brittleness inherent in traditional neuro-symbolic systems, establishing a new ceiling for geometry problem-solving performance.

\vspace{-0.25cm}
\subsection{Iteration Rounds}
\begin{figure}[h]  
\vspace{-0.2cm}
    \centering
    \includegraphics[width=1.0\linewidth]{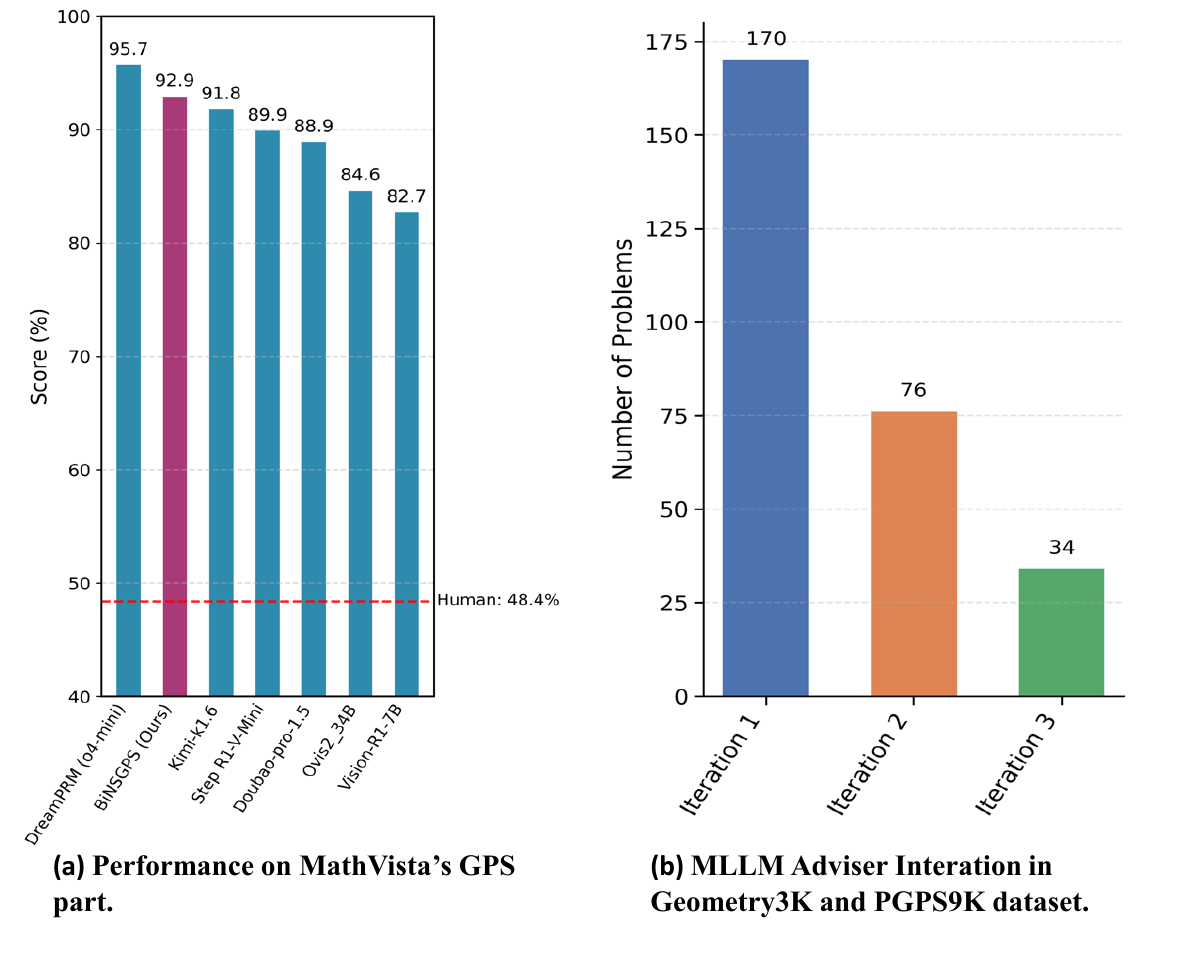}
    \caption{Performance in MathVista (a) and Adviser Interation in Geometry3K and PGPS9K dataset (b).}
    \label{fig:mathvista and iteratioin}
\vspace{-0.4cm}
\end{figure}

\vspace{-0.1cm}
To evaluate the efficiency of the Bidirectional Neuro-Symbolic Interaction (BiNS), we analyze the number of rounds required by the MLLM Adviser to reach a valid solution. In our framework, we implement a safety threshold with a maximum iteration limit of $T=3$.
The empirical distribution of success rounds using Qwen3-VL-Plus as MLLM Adviser, as illustrated in Fig.~\ref{fig:mathvista and iteratioin}(b), demonstrates the rapid convergence of the BiNS loop. Among the cases requiring intervention beyond the initial formalization, the system successfully resolves 171 problems in a single iteration, 77 cases in two iterations, and 35 cases in three iterations(Fallback).

The high early success rate reflects the precision of the Symbolic Solver’s diagnostic feedback. By identifying exact symbolic conflicts, it enables targeted corrections instead of stochastic search. Few cases require a third iteration, confirming efficient mastery of complex reasoning without excessive computation. This validates BiNSGPS as a practical geometry-solving paradigm, balancing multimodal intuition with rapid symbolic verification.

\vspace{-0.2cm}
\section{Conclusion}
\vspace{-0.2cm}
This paper introduces BiNSGPS, a framework that reformulates the neuro-symbolic architecture through Bidirectional Neuro-Symbolic Interaction (BiNS). By establishing a closed-loop feedback mechanism between an MLLM Adviser and a Symbolic Solver, we enables the system to iteratively rectify formal representations and propose auxiliary hypotheses, effectively synergizing the adaptive intuition of neural models with the mathematical rigor of symbolic reasoning.

\section*{Limitations}
While BiNSGPS achieves state-of-the-art performance on Geometry3K and PGPS9K, we identify the automated generation of auxiliary constructions as a key frontier for the next iteration of neuro-symbolic reasoning.
Currently, the Geometry3K and PGPS9K datasets impose minimal requirements for auxiliary points and lines. Consequently, we deliberately refrain from prompting the MLLM to generate such elements to mitigate potential hallucinations. 
In addition, although we use prompts to control the output of MLLMs, in certain cases, MLLMs may lose control and repeatedly doubt their previous reasoning steps, leading to excessively long outputs. Further research is needed to enhance the controllability of MLLMs.


\bibliography{custom}

\clearpage
\appendix

\section{GPS Problem Formulation}
The task of Geometry Problem Solving (GPS) is formally defined as follows. The system is provided with a dataset $\{\mathcal{D}_i, \mathcal{T}_i\}$, where $\mathcal{D}_i$ denotes the geometric diagram, and $\mathcal{T}_i$ denotes the corresponding textual description (e.g., "Find the measure of angle ABC"). Given both the diagram and the text, the system is required to generate the corresponding answer $A_i$, along with the associated solution steps $\mathcal{S} = \{s_1, s_2, \dots, s_n\}$. 
We evaluate the system under two primary settings. In the competition mode, problems are presented in an open-ended format where the system must directly output the final value. In the choice mode, problems are presented in a multiple-choice format, requiring the system to select the correct option from four candidates $\{\mathcal{D}_i, \mathcal{T}_i\}$.

\section{Experiment details}
\subsection{Computational Cost Analysis}

To rigorously evaluate the cost introduced by our proposed framework, we conducted a detailed cost analysis on the Geometry3K and PGPS9K datasets. Our analysis reveals that the BiNS framework incurs an average computational cost of 2,634 tokens per question. This token consumption is deemed justifiable when weighed against the substantial gains in accuracy. Concurrently, the BiNSGPS framework introduces an average latency of 3.54 minutes; this temporal overhead falls well within acceptable bounds for practical deployment.

\subsection{Choice Mode}
In many existing geometry solvers, the presence of multiple-choice options is used to simplify the problem space via elimination or "guess-and-check" heuristics. In contrast, BiNSGPS maintains a "completion-first" philosophy, where the choices serve exclusively as a Consistency Check within the bidirectional loop.

The transition from standard deduction to Choice-based rectification follows a strict logical sequence: The Symbolic Solver first attempts to derive a solution $S_{sym}$ using only the formal representations $\mathcal{L}$ and theorems $\mathcal{R}$, without access to the candidate options $\{A, B, C, D\}$. 
If $S_{sym}$ does not align with any of the candidate choices, the system identifies a symbolic-perceptual mismatch. This discrepancy acts as the trigger for the MLLM Adviser. The Adviser is then provided with the candidate choices as auxiliary constraints.

A primary advantage of this approach is its ability to prevent the system from "guessing" based on visual intuition or heuristic shortcuts. While pure MLLMs often select an option based on plausible, BiNSGPS instead treats the choice set as a high-level diagnostic signal to re-examine the answer. This ensures that the final selection is not merely a statistical probability but is always supported by a valid symbolic proof chain derived through the Solver.

\subsection{Fallback rate and Accuracy}
To clarify the influence of fallback mode, we analyse the trigger frequency and accuracy of fallback mode. Across the two datasets, only 105 questions triggered BiNSGPS's fallback mode, accounting for just 6.56\% of the datasets. And the accuracy in fallback mode is 33.3\%. This accuracy is largely due to the questions that can't be solved in normal mode are the hardest part of the datasets and MLLM Adviser can hardly handle them, the accuracy of Qwen3-VL-Plus in these question is 29.5\%, also largely lower than the average level. But as stated in the paper, the fallback mode mainly serves as a robustness safeguard, ensuring that BiNSGPS can still return an answer instead of collapsing without output.

\subsection{PGDPNet}
The reliability of a neuro-symbolic system depends heavily on the accuracy of its initial perception layer. In BiNSGPS, we utilize PGDPNet as our specialist neural network for diagram parsing to ensure the initial formal representation set $\mathcal{L}$ is mathematically grounded. To justify its role as a robust "anchor" for our experiments, we provide its performance metrics on geometric primitive detection and spatial relationship extraction below.

As summarized in Table~\ref{tab:pgdp_stats}, PGDPNet achieves near-perfect scores across all primary geometric primitives.

\begin{table}[h]
\centering
\caption{PGDPNet Precision, Recall, and F1 Score for Primitive Detection}
\label{tab:pgdp_stats}
\resizebox{0.98\columnwidth}{!}{
\begin{tabular}{|l|c|c|c|}
\hline
\textbf{Primitive} & \textbf{Precision (\%)} & \textbf{Recall (\%)} & \textbf{F1 Score (\%)} \\
\hline
Point & 99.65 & 99.71 & 99.68 \\
Line & 99.30 & 99.51 & 99.40 \\
Circle & 99.85 & 99.96 & 99.90 \\
\hline
\end{tabular}
}
\end{table}

Beyond identifying isolated primitives, PGDPNet is tasked with extracting topological relations such as tangency, collinearity, and intersections. For these complex positional relationships, the model achieves a Precision of 99.13\%, a Recall of 98.60\%, and an F1 Score of 98.86\%.

These metrics demonstrate that PGDPNet provides an exceptionally sound foundation for the BiNSGPS pipeline. By maintaining $99\%+$ accuracy in primitive detection, the network minimizes the "noise" typically introduced by general-purpose MLLMs. 

\vspace{-0.2cm}
\section{Symbolic Solver}
\label{appendix:Symbolic Solver}
\subsection{Theorems Apply Strategy}
To maximize problem-solving performance, the Symbolic Solver employs a complexity-ascending heuristic that orchestrates theorem application strictly according to the cardinality of their required conditions. This strategic ordering is paramount for system efficacy; given that deeper conclusions frequently depend on the derivation of intermediate results, an ill-sequenced application order can precipitate premature termination, thereby preventing the system from converging upon the final solution within the stipulated iteration limit.

\subsection{The Minimal Solution Graph}
Minimal solution graph is the reference for proof generation. Thanks to the hypergraph-based solving process, Solver can track the trace of prove. The Symbolic Solver performs a recursive back-trace from the destination node to its fundamental root nodes within the hypergraph. 
In hypergraph expansion, every node except the initial ones records the theorems and predecessor nodes utilized during reasoning. By backtracking from the final node, the Symbolic Solver can trace all necessary paths from the root to the target node, thereby enabling the identification of the Minimal Solution Graph.
This extraction identifies the minimal solution graph, filtering out redundant derivations to ensure a concise and step-wise logical path.

\subsection{Rule-based Filter}
To ensure the output remains within the permitted logic space of the symbolic engine, BiNSGPS utilizes a Rule-Based Filter that acts as a deterministic security layer. This filter provides a lightweight yet robust mechanism to validate representations before send to solver.

The filter permits Algebraic and Trigonometric Rules. It validates Basic Elements as defined in the primary representation space and accepts Fundamental Operators such as Equals and Find. Quantitative Attributes are strictly limited to AreaOf, RadiusOf, MeasureOf, and LengthOf. Finally, Geometric Relations are restricted to a high-fidelity subset: Parallel, Perpendicular, IsSimilarity, IsCongruent, Tangent, IsMidpointOf, and IsMidsegmentOf.

In the Multimodal Representations Alignment stage, if a non-compliant or hallucinated assertion is detected, the checker provides immediate feedback, prompting the MLLM to regenerate the representation. This multi-layered verification transforms each problem ($\mathcal{D}_i$, $\mathcal{T}_i$) into a logically consistent initial representation set $\mathcal{L}_i$, ready for Symbolic Solver's rigorous deduction.

We also implement a rule-based filter that acts as a syntactic gateway between the MLLM Adviser and the Symbolic Solver. This filter automatically screens and excludes any generated assertions that fall outside the solver’s permissible assertion space, effectively neutralizing the risk of format-induced failures. It also verifies that every referenced point and circle corresponds to an entity already identified in the initial formalization, preventing the MLLM from hallucinating non-existent vertices or primitives.

By operating as a rigid pattern-matching gateway, the filter ensures that the BiNSGPS remains mathematically sound, protecting the Symbolic Solver from the inherent stochasticity of neural model outputs.

\subsection{Formal Representation Definitions}
To ensure a rigorous interface between the MLLM and the symbolic engine, BiNSGPS utilizes a structured representation space. Following the principles of AutoGPS, we categorize these logic forms into two primary groups: basic structural foundations and high-level geometric constraints.

The first Table~\ref{tab:predicates} comprises Basic Elements and Algebraic Rules, which define the existence of geometric primitives (Points, Lines, Circles, Polygons) and the fundamental arithmetic operations used to model their numerical relationships. The second Table~\ref{tab:predicates2} consists of Relations and Quantitative Attributes, which describe how these primitives interact (e.g., parallelism, tangency, or collinearity) and define their measurable properties such as length, angle measure, and area.

By mapping multimodal inputs to this standardized space, the framework ensures that every step taken by the Symbolic Solver is grounded in a formal axiomatic system. This structured approach allows the MLLM Adviser to perform rectifications by targeting specific predicates in the logic set.

\vspace{-0.2cm}
\section{Feedback mode}
We acknowledge that AutoGPS also incorporates feedback, but its mechanism is fundamentally performs one-time, pre-reasoning validation of formalized representations: once the initial representation passes the self-consistency check, the loop terminates permanently. Consequently, even if the formalization is initially deemed consistent, hidden contradictions may still emerge during deduction, and AutoGPS has no mechanism to recover. In contrast, BiNSGPS establishes a runtime cognitive collaboration: when the symbolic solver encounters conflicts or deadlocks during reasoning, the MLLM Adviser intervenes to perform diagnostic correction guided by derived contradictions, or to propose auxiliary theorems that serve as new deductive steps to resolve the impasse.
We elevate the MLLM from a one-time validator to a collaborative partner throughout the entire proof search, and this architectural shift is what drives the substantial performance gain.

\begin{figure*}[th]
  \centering
  \includegraphics[width=\linewidth]{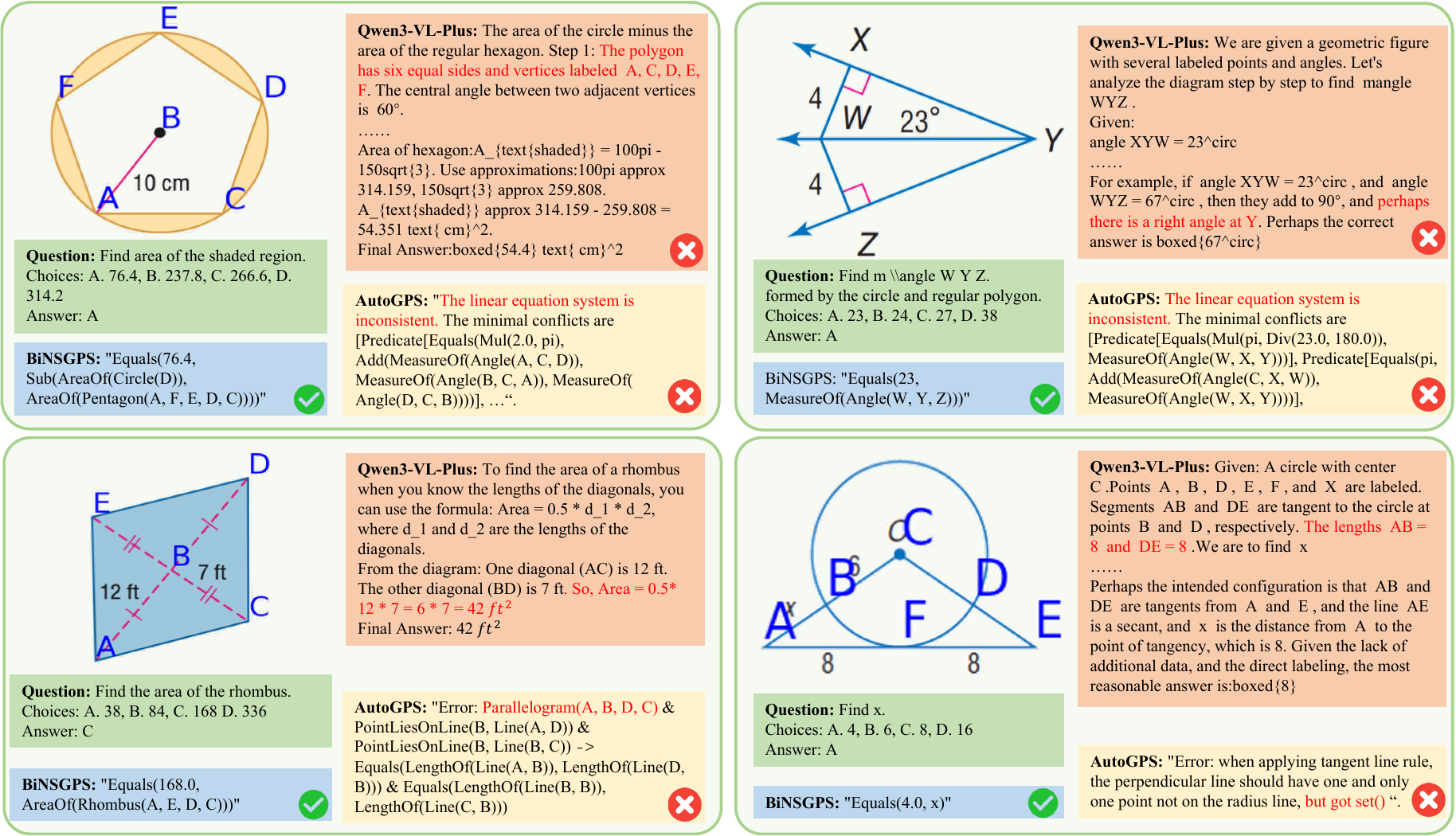}
  \caption{Failure cases of current methods. Qwen3-VL-Plus exhibits hallucination induced errors during reasoning(up-left: recognize pentagon as hexagon), producing an incorrect conclusion. The neural-symbolic method (Auto-GPS) will immediately fail when the logic forms have conflict. While BiNSGPS show more robustness with MLLM's feedback. \textcolor{red}{Red} annotations indicate correct/erroneous reasoning or answers.}
  \label{fig:case study}
\end{figure*}

\section{Case Study}
To provide a more comprehensive illustration of the efficacy and limitations of our proposed method, we present a series of case studies herein.

\vspace{-0.1cm}
\subsection{Illustrative Case}
We present illustrative case studies in Fig.~\ref{fig:case study} to demonstrate the BiNSGPS's performance on diverse geometry problems and show its advantages over other methods.

\vspace{-0.1cm}
\subsection{Success Case}
\begin{figure*}[h]  
    \centering
    \includegraphics[width=\linewidth]{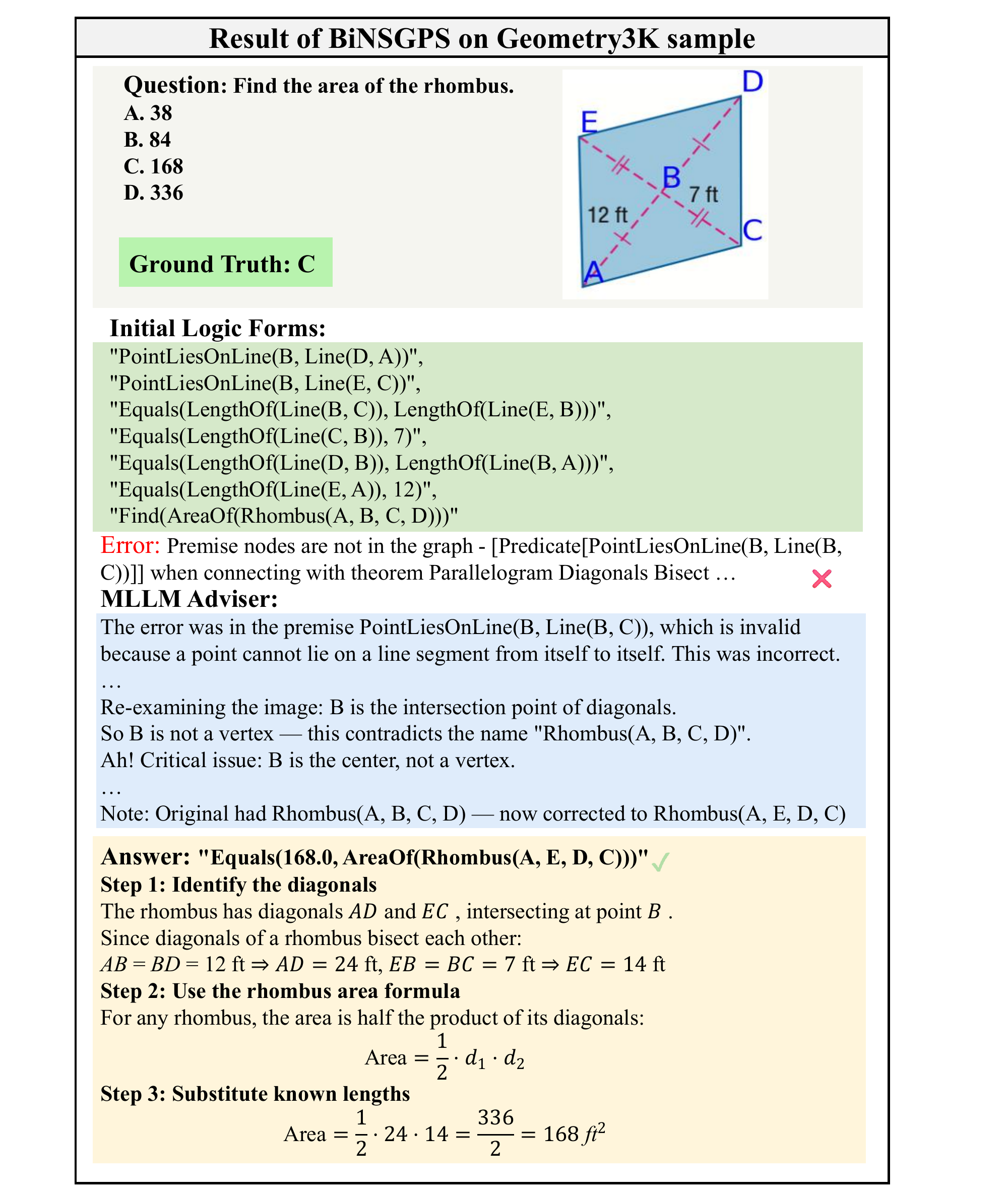}
    \caption{Success case of BiNSGPS's Rectify Inconsistent Representations.}
    \label{fig:success_case}
\end{figure*}

\begin{figure*}[h]  
    \centering
    \includegraphics[width=\linewidth]{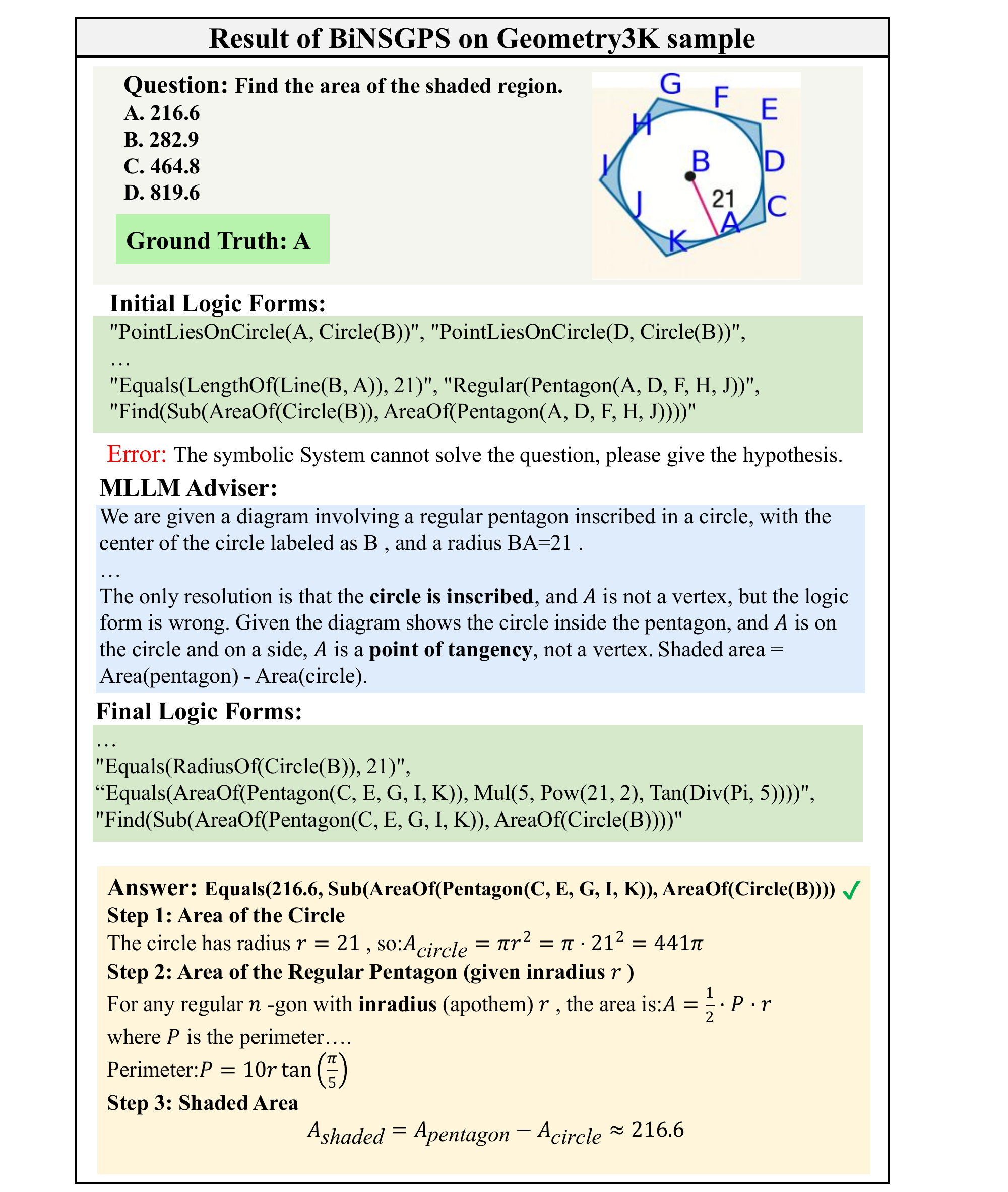}
    \caption{Success case of BiNSGPS's Propose Auxiliary Hypotheses.}
    \label{fig:success_case2}
\end{figure*}

Here we present detailed example of BiNSGPS's success case to illustrate our framework's effect. The success of BiNSGPS is defined by its ability to resolve representational conflicts and leverage symbolic feedback. The system successfully recovers from initial formalization errors and overcomes limitation by proposing auxiliary hypotheses as shown in Fig.~\ref{fig:success_case} and Fig.~\ref{fig:success_case2}


\vspace{-0.1cm}
\subsection{Fail Cases}
\begin{figure*}[h]  
    \centering
    \includegraphics[width=\linewidth]{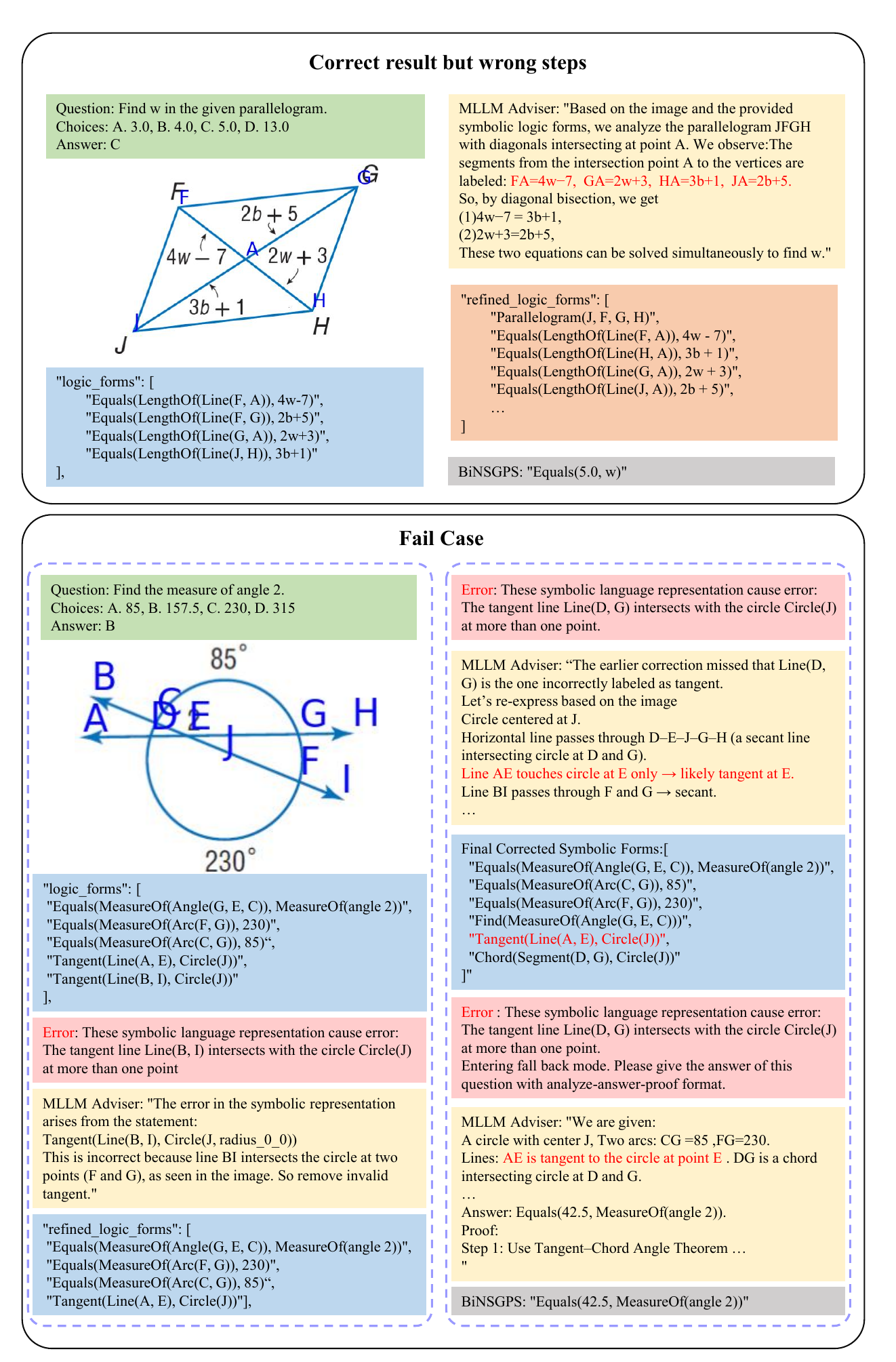}
    \caption{Fail cases of BiNSGPS. Including correct results but wrong steps case (up) and fail case (bottom)}
    \label{fig:fail_case}
\end{figure*}

Here we present the failure cases of BiNSGPS as shown in Fig.~\ref{fig:fail_case}.
For the correct results but wrong steps, MLLM Adviser fails to assign value to corresponding element but drives the right answer in coincidence.
For the fail case, the diagram usually contains too many element that MLLM Adviser can hardly detect them, and leads to failure.

\begin{table*}[h]
\centering
\caption{Predicate and literal definitions for the formal representations (1).}
\label{tab:predicates}
\begin{tabular}{ll}
\toprule
\textbf{Literals} & \textbf{Explanation} \\
\midrule
Line(A,B) & A line segment with endpoints A and B \\
Angle(A) & The angle with point A as vertex \\
Angle(A,B,C) & Angle ABC with B as the vertex \\
Triangle(A,B,C) & Triangle with vertices A, B, and C \\
Quadrilateral(A,B,C,D) & Quadrilateral with vertices A, B, C, and D \\
Parallelogram(A,B,C,D) & Parallelogram with vertices A, B, C, and D \\
Square(A,B,C,D) & Square with vertices A, B, C, and D \\
Rectangle(A,B,C,D) & Rectangle with vertices A, B, C, and D \\
Rhombus(A,B,C,D) & Rhombus with vertices A, B, C, and D \\
Trapezoid(A,B,C,D) & Trapezoid with vertices A, B, C, and D \\
Kite(A,B,C,D) & Kite with vertices A, B, C, and D \\
Polygon(A,B,C,...) & Polygon with vertices A, B, C, etc. \\
Pentagon(A,B,C,D,E) & Pentagon with vertices A, B, C, D, and E \\
Hexagon(A,B,C,D,E,F) & Hexagon with vertices A, B, C, D, E, and F \\
Octagon(A,B,C,D,E,F,G,H) & Octagon with vertices A, B, C, D, E, F, G, and H \\
Circle(A) & Circle with center A \\
Circle(O, r) & Circle with center O and radius r \\
Arc(A,B) & Minor arc with A and B as endpoints on circle \\
Arc(A,B,C) & Arc that passes through points A, B, and C \\
Sector(O,A,B) & Sector of a circle with center O \\
SinOf(var) & Sine of var \\
CosOf(var) & Cosine of var \\
TanOf(var) & Tangent of var \\
CotOf(var) & Cotangent of var \\
HalfOf(var) & Half of var \\
SqrtOf(var) & Square root of var \\
RatioOf(var1,var2) & Ratio of var1 to var2 \\
Add(var1,var2,...) & Addition of var1, var2 ... \\
Mul(var1,var2,...) & Multiplication of var1, var2, ... \\
Sub(var1,var2) & Subtraction of var2 from var1 \\
Div(var1,var2) & Division of var1 by var2 \\
Pow(var1,var2) & var1 raised to the power of var2 \\
Equals(var1,var2) & var1 equals var2 \\
Find(var) & Find the value of the variable \\
\bottomrule
\end{tabular}
\end{table*}

\begin{table*}[h]
\centering
\caption{Predicate and literal definitions for the formal representations (2).}
\label{tab:predicates2}
\begin{tabularx}{\textwidth}{l X}
\toprule
\textbf{Literals} & \textbf{Explanation} \\
\midrule
Equilateral(Polygon(A,B,C,D)) & Polygon ABCD is equilateral \\
Regular(Polygon(A,B,C,D)) & Polygon ABCD is regular \\
AreaOf(Shape(A)) & Area of the Shape A \\
PerimeterOf(Shape(A)) & Perimeter of the Shape A \\
RadiusOf(Circle(O)) & Radius of the circle O \\
DiameterOf(Circle(O)) & Diameter of the circle O \\
CircumferenceOf(Circle(O)) & Circumference of the circle O \\
MeasureOf(Angle(A, B, C)) & Measure of the angle ABC \\
MeasureOf(Arc(A, B)) & Measure of the arc AB \\
LengthOf(Line(A, B)) & Length of the line segment AB \\
PointLiesOnLine(A,Line(B,C)) & Point A lies on Line BC \\
PointLiesOnCircle(A,Circle(O)) & Point A lies on the circle O \\
Parallel(Line(A,B),Line(C,D)) & Line AB is parallel to Line CD \\
Perpendicular(Line(A,B),Line(C,D)) & Line AB is perpendicular to Line CD \\
BisectsAngle(Line(A,B),Angle(X,A,Y)) & Line AB bisects angle XAY \\
Congruent(Triangle(A,B,C),Triangle(D,E,F)) & Triangle ABC is congruent to triangle DEF \\
Similar(Triangle(A,B,C),Triangle(D,E,F)) & Triangle ABC is similar to triangle DEF \\
Tangent(Line(A,B),Circle(O)) & Line AB is tangent to circle O \\
CircumscribedTo(Shape(A),Shape(B)) & Shape A is circumscribed to shape B \\
InscribedIn(Shape(A),Shape(B)) & Shape A is inscribed in the shape B \\
IsMidpointOf(C,Line(A,B)) & Point C is the midpoint of line AB \\
IsCentroidOf(O,Triangle(A,B,C)) & Point O is the centroid of triangle ABC \\
IsIncenterOf(O,Triangle(A,B,C)) & Point O is the incenter of triangle ABC \\
IsRadiusOf(Line(O,A),Circle(O)) & Line OA is a radius of circle O \\
IsDiameterOf(Line(A,B),Circle(O)) & Line AB is a diameter of circle O \\
IsMidsegmentOf(Line(A,B),Triangle(D,E,F)) & Line AB is a midsegment of triangle DEF \\
IsPerpendicularBisectorOf(Line(A,B),Line(C,D)) & Line AB is the perpendicular bisector of line CD \\
IsMedianOf(Line(E,F),Trapezoid(A,B,C,D)) & Line EF is the median of trapezoid ABCD \\
IsMedianOf(Line(E,F),Triangle(A,B,C)) & Line EF is a median of triangle ABC \\
\bottomrule
\end{tabularx}
\end{table*}
\clearpage

\end{document}